\pdfoutput=1

\documentclass[11pt]{article}

\usepackage[preprint]{acl}

\usepackage{times}
\usepackage{latexsym}

\usepackage[T1]{fontenc}

\usepackage[utf8]{inputenc}

\usepackage{microtype}

\usepackage{inconsolata}

\usepackage{graphicx}

\usepackage{tabularx}
\usepackage{tabularray}

\usepackage[colorinlistoftodos,textsize=tiny,disable]{todonotes}

\usepackage{amsmath}

\usepackage{enumitem}

\usepackage{listings}

\usepackage{tcolorbox}
\usepackage{caption}

\title{Can LLMs Extract Frame-Semantic Arguments?}

\author{Jacob Devasier, Rishabh Mediratta, Chengkai Li \\
        University of Texas at Arlington \\ cli@uta.edu}

\begin{document}
\maketitle
\begin{abstract}
Frame-semantic parsing is a critical task in natural language understanding, yet the ability of large language models (LLMs) to extract frame-semantic arguments remains underexplored. This paper presents a comprehensive evaluation of LLMs on frame-semantic argument identification, analyzing the impact of input representation formats, model architectures, and generalization to unseen and out-of-domain samples. Our experiments, spanning models from 0.5B to 78B parameters, reveal that JSON-based representations significantly enhance performance, and while larger models generally perform better, smaller models can achieve competitive results through fine-tuning. We also introduce a novel approach to frame identification leveraging predicted frame elements, achieving state-of-the-art performance on ambiguous targets. Despite strong generalization capabilities, our analysis finds that LLMs still struggle with out-of-domain data. 
\end{abstract}

\section{Introduction}
\begin{figure}
  \centering
  \includegraphics[width=\linewidth, trim={0 7mm 0 10mm}]{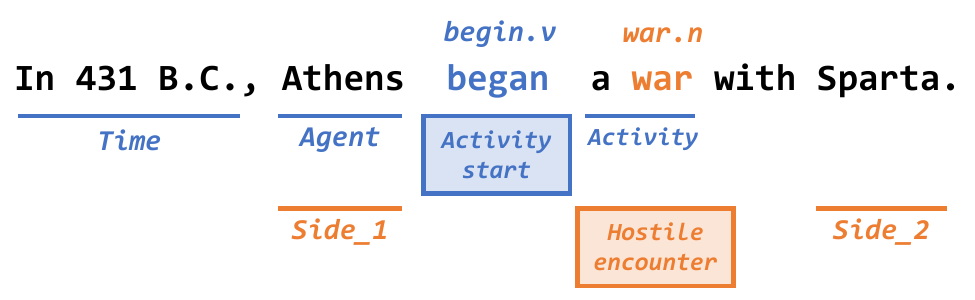}
  
  \caption{\label{fig:fsp-example} An example of frame-semantic annotations.}
\end{figure}

Frame-semantic parsing~\cite{gildea-jurafsky-2002-automatic} is a fundamental task in natural language understanding that involves identifying semantic frames~\cite{baker-etal-1998-berkeley-framenet} and their associated elements within a sentence. This process is typically divided into three sub-tasks: target identification (detecting words that evoke frames, e.g., began in Figure~\ref{fig:fsp-example}), frame identification (determining the specific frame evoked, i.e., Activity\_start), and argument identification (extracting frame elements, i.e., Time, Agent, and Activity).

Traditional approaches to frame-semantic parsing have found success with supervised classification models~\cite{chakma2024semantic}. However, the potential of large language models (LLMs) for this task remains largely unexplored. Recent work~\cite{Su2024} has applied in-context learning with LLMs but found their performance to be significantly weaker, raising concerns about their ability to accurately extract frame-semantic arguments.

In this work, we conduct a comprehensive study on the effectiveness of LLMs for frame-semantic argument identification, evaluating key factors that may influence performance, including input representation formats, model architecture and scale, and generalization to unseen and out-of-domain samples. Our experiments span a diverse range of state-of-the-art LLMs, from 0.5B-parameter models to 78B-parameter models, including both open-source models (e.g., Qwen 2.5~\cite{qwen2025qwen25technicalreport}, Llama 3) and closed-source models (e.g., GPT-4o).

Recent work~\cite{devasier-etal-2024-robust} has explored unifying target identification and frame identification by applying frame identification models to candidate targets. We also expand on this idea with a novel method for unifying frame identification and argument identification by leveraging predicted frame elements of candidate frames. 

Our findings reveal several important insights. First, we confirm that LLMs struggle in zero-shot and few-shot settings, reinforcing prior concerns about their reliability for frame-semantic parsing. Second, we demonstrate that the choice of input representation significantly impacts model performance, with JSON-based formats showing superior results. Surprisingly, we found that while model scale generally correlates with better performance, smaller models like Qwen 2.5 (3B) outperform the much larger Llama 3.3 (70B). Finally, our proposed frame identification method using predicted frame elements achieves competitive performance, particularly for ambiguous targets (i.e., words that can evoke multiple frames), where it surpasses previous state-of-the-art approaches by +1.2\%.

To summarize, this work makes the following contributions.
\begin{itemize}[noitemsep,topsep=0pt,leftmargin=*]
    \item We conduct a systematic evaluation\footnote{All of our training and evaluation code is available at \url{https://anonymous.4open.science/r/llm-fsp-831F}} of different input/output representations for frame-semantic parsing with generative LLMs.
    \item We produce comprehensive benchmarks of different LLM architectures at varying scales, resulting in a +3.9\% F1 score improvement over the previous best argument identification model. 
    \item We developed a novel frame identification approach leveraging predicted frame elements on candidate frames which achieves state-of-the-art performance on ambiguous targets.
\end{itemize}

\section{Background and Related Works}

Frame-semantic parsing is the automatic extraction of semantic frames and their elements. The task is often applied to FrameNet, a large corpus of frame-semantic annotations and definitions, and is typically separated into three subtasks: target identification, frame identification, and argument extraction (sometimes referred to as frame-semantic role labeling). Target identification is the process of identifying targets--instances of predefined lexical units--in a sentence. Lexical units are unique pairings of words and their meaning, indicated in FrameNet using the word's lemma and part-of-speech (e.g., \textit{begin.v} and \textit{war.n} in Figure~\ref{fig:fsp-example}) which are associated with a particular frame. Frame identification is the process of identifying the frames evoked in a sentence (e.g., Activity\_start and Hostile\_encounter in Figure~\ref{fig:fsp-example}), often done by classifying the previously extracted targets. Argument extraction is the process of extracting all frame elements of a particular frame evoked in a sentence (e.g., Time, Agent, and Activity for the Activity\_start frame in Figure~\ref{fig:fsp-example}). 

Nearly all previous systems use classification methods for argument extraction~\cite{chakma2024semantic}. These approaches are primarily dominated by BERT-like encoder models. Argument extraction is often structured as either a token or segment classification task~\cite{Su2024, aged2023, zheng-etal-2022-double, Bastianelli2020EncodingSC, opensesame, lin-etal-2021-graph} or a span identification task~\cite{Ai_Tu_2024, devasier-etal-2024-claimlens, aged2023, chen-etal-2021-joint}. Token/segment classification approaches classify each token or sequence of tokens as one of the frame elements, span identification approaches identify the beginning and end positions of each frame element. One previous study on argument extraction has used zero/few shot in-context learning with a simple prompt on Llama 2~\cite{Su2024}, but observed very poor performance. Another previous study on the very similar task of semantic role labeling has also found a large performance discrepancy when using LLMs~\cite{limitations-of-llms-ning-2024}.

\section{Methodology}
\begin{table*}
\centering
\begin{tblr}{|Q[l]|Q[4.5cm, c]|Q[8cm, l]|}
\hline
\textbf{Representation} & \textbf{Input} & \textbf{Output} \\
\hline
Markdown & \SetCell[r=4]{c}{Your **contribution** to Goodwill will mean more than you may know.} & - Donor: Your \linebreak - Recipient: to Goodwill\\
\hline
XML Tags &  & \texttt{<Donor>Your</Donor> contribution <Recipient>to Goodwill</Recipient> will mean more than you may know.} \\
\hline
JSON-Existing &  & \texttt{\{``Donor": ``Your", ``Recipient": ``to Goodwill"\}} \\
\hline
JSON-Complete &  & \texttt{\{``Donor": ``Your", ``Recipient": ``to Goodwill", ``Theme": ``", ``Place": ``", ...\}} \\
\hline
\end{tblr}
\caption{\label{tab:fe-representation}Representation formats for the given input and outputs.}
\end{table*}

\subsection{Input Representation Design}
\label{sec:input-representation}
Previous research has shown that large language models are sensitive to input formatting~\cite{Sclar2023QuantifyingLM} and that different representations can result in different model performance~\cite{tam-etal-2024-speak,textsql-eval-gao-2024,exploring-marcos-2024}. To study these effects on frame-semantics, we systematically evaluated multiple input-output representation formats to determine their impact on frame element extraction performance.

For all input formats, we wrap the target word or phrase in double asterisks, as shown in Table~\ref{tab:fe-representation}, to explicitly mark the token that evokes the frame. This marking helps focus the model's attention on the relevant part of the sentence when making frame element predictions, ensuring that the model identifies frame elements for the correct target.

We developed and tested four distinct representation formats. The Markdown format offers a simple, human-readable approach where frame elements are represented as a markdown list. Each list item contains a frame element name paired with its corresponding text span from the sentence. This format only includes frame elements that the model predicts are present in the input. The XML Tags format provides a structured approach that uses XML-style tags to wrap frame elements within the sentence text. The tag names correspond to frame element names, providing both semantic labeling and precise positional information without requiring additional processing.

We also developed two JSON-based formats. The JSON-Existing format uses frame element names as keys and their corresponding text spans from the sentence as values. Similar to the Markdown format, this only includes predicted frame elements. The JSON-Complete format provides an exhaustive representation different from previous representations that includes all possible frame elements as keys, with empty strings as values for elements not found in the sentence. This format was designed to test whether explicitly presenting all possible frame elements might improve model performance. Examples of each representation format are provided in Table~\ref{tab:fe-representation}, illustrating how they encode the same semantic information in different ways.

\subsection{Model Selection and Implementation}
To ensure a comprehensive evaluation across the current LLM landscape, we selected models varying in size, architecture, and accessibility. Our selection criteria focused on three key dimensions. In terms of model scale, we included models ranging from 0.5B to 78B parameters, categorizing them into small-scale (0-14B parameters) and large-scale (14B+ parameters) groups to analyze the impact of model size on performance. For architecture diversity, we selected top-performing models from the HuggingFace LLM leaderboard, with particular focus on Qwen 2.5 and Llama 3.2, which have shown strong performance on various tasks.

We included both open-source models (Qwen 2.5, Llama 3, and Phi-4) and closed-source systems (GPT-4o and GPT-4o-mini) to compare performance across different levels of model accessibility. For the open-source models, we implemented fine-tuning using LoRA~\cite{hu2021loralowrankadaptationlarge}. We used $r=16$ for all models except Llama 3.3 and Qwen 2.5 (72B) where we used $r=32$, according to best practices. This approach allowed us to optimize model performance while maintaining reasonable computational requirements.

\subsection{Evaluation}
Our evaluation framework was designed to comprehensively assess model performance across different scenarios and conditions. We began by testing each representation's effectiveness using controlled experiments with GPT-4o-mini. Model performance was evaluated using standard metrics including precision, recall, F1 score, and accuracy with exact match criteria.

To understand data requirements and efficiency, we analyzed performance with varying amounts of training data to understand data efficiency and saturation points. We also conducted extensive testing of model performance on unseen frames, unseen frame elements, and out-of-domain samples. Finally, we analyzed the distribution of argument extraction performance for each frame to gain a granular understanding. This evaluation framework enables us to systematically evaluate LLMs' capabilities in frame-semantic parsing while providing insights into the impact of different design choices and implementation strategies.

\section{Experiments}
In this section, we thoroughly evaluate the performance of LLMs on frame-semantic parsing through several experiments designed to address three primary research questions: RQ1) How does the representation of FEs impact performance? RQ2) How does model architecture and scale impact performance? RQ3) Are LLMs better on out-of-domain/unseen samples than previous non-LLM methods?

\subsection{Dataset}
We utilize the FrameNet 1.7~\cite{FramenetExtended} dataset for our primary experiments. FrameNet provides detailed definitions of semantic frames and their elements, including partially-annotated exemplar sentences for each frame and a corpus of fully-annotated sentences (referred to as "full-text annotations"). We use the full-text annotations for model training due to their complete coverage of frames and frame elements.

Following established conventions from \citet{opensesame} and \citet{das-smith-2011-semi}, we use standard train/test splits of non-overlapping documents. The training split consists of 3,353 sentences which evoke 19,391 frames with 34,219 frame elements, while the test split contains 1,247 sentences evoking 6,714 frames and 11,302 frame elements. For out-of-domain evaluation, we use the YAGS dataset, which contains 2,093 test sentences evoking 364 frames with 4,162 frame elements.

\begin{table}
    \centering
    \begin{tabular}{lcccc}
        \hline
        \textbf{Format} & \textbf{P} & \textbf{R} & \textbf{F1} & \textbf{Acc} \\ 
        \hline
        XML-tag & 0.318 & 0.368 & 0.342 & 0.206 \\ 
        JSON-all & 0.356 & \textbf{0.577} & 0.440 & 0.282 \\ 
        Markdown & 0.376 & 0.554 & 0.448 & 0.289 \\ 
        JSON-exist & \textbf{0.416} & 0.543 & \textbf{0.471} & \textbf{0.308} \\ 
        \hline
    \end{tabular}
    \caption{Few-shot performance metrics for different FE representations using GPT-4o-mini with 0 temperature.}
    \label{tab:representation_performance}
\end{table}


\subsection{Frame Element Representations (RQ1)}
To answer RQ1, we evaluate various FE representation approaches using in-context learning with GPT-4o-mini. Our experiments (Table~\ref{tab:representation_performance}) reveal that JSON-Existing achieves superior performance with significant margins in precision (+4\%), F1 score (+3.4\%), and accuracy (+1.9\%). While JSON-Complete showed higher recall (+4.4\%), we attribute this to the simplified cognitive load of outputting all possible frame elements rather than selecting relevant ones. This comprehensive approach leads to more complete but less precise predictions. Notably, XML tag representation performed significantly worse, showing a 12.9 percentage point reduction in F1 score compared to JSON-exist, likely due to the difficulty of the representation. This also suggests that FrameNet's annotations may not be included in common pre-training data as they are originally in XML format. 

We found that using JSON-Existing results in the highest precision, F1 score, and accuracy, by significant margins (+4\%, +3.4\%, and +1.9\%, respectively). Interestingly, JSON-Complete had a higher recall (+4.4\%). We believe this is due to a reduced cognitive load given by instructing the LLM to output each frame element instead of just the ones which exist in the sentence. This likely reducing the chance of missing particular frame elements, resulting in higher recall. We also found that XML tags performed the worst by a large margin, likely due to the added positional requirements introduced. This indicates that FrameNet is likely not included in the pretraining corpus of LLMs as its native annotations are in XML format. 

\subsection{Generating LLM Instructions}
To validate our instruction creation process, we conducted a comparative study using instructions generated by GPT-4o. The automated approach included all frame-specific information and examples to allow flexibility in prompt generation. Despite being similar to our manual instructions (ROUGE-1/L score: 0.59/0.36), the automated instructions resulted in significantly lower performance (F1 score: 0.225 vs. 0.471). ~\todo{should we clarify that we use the diff prompts but the same llm to get these numbers?} We found that this was primarily due to the LLM predicting frame elements that do not exist, leading us to proceed with manually-crafted instructions for subsequent experiments. 

\lstset{
     basicstyle=\footnotesize\scriptsize,
    backgroundcolor=\color{gray!10}, 
    linewidth=\columnwidth,
    breaklines=true,
    frame=single, 
    rulecolor=\color{black}, 
    showstringspaces=false, 
    numbers=left, 
    numberstyle=\tiny\color{gray}, 
    xleftmargin=0em, 
    numbers=none, 
    moredelim=**[is][\color{gray!70}]{@}{@} 
}
\begin{lstlisting}[caption={Sample prompt used for zero-shot evaluation.}, label=lst:prompt]
### Task:
You are given a sentence and a frame with its associated frame elements and sometimes examples. Your task is to label the frame elements in the sentence using JSON. Keys should only be one of the defined frame elements. Do not make up your own frame elements, and do not remove or change the input in any way. Identify the frame elements based on the highlighted target word. 

### Frame Information:
Frame Name: Awareness
Frame Definition: A Cognizer has a piece of Content in their model of the world. ... [omitted for brevity] ...
Examples:
  - Your boss is aware of your commitment. -> {"Cognizer": "Your boss", ...}
  ... [omitted] ...

Frame Elements:
Cognizer (Core): The Cognizer is the person whose awareness of phenomena is at question. 
  - Your boss is **aware** of your commitment. -> {"Cognizer": "Your boss"}
  ... [omitted] ...

Explanation (Extra-Thematic): The reason why or how it came to be that the Cognizer has awareness of the Topic or Content.

### Notes:
- Return the tagged sentence in a ```json ``` code block.
- Texts must not overlap.

\end{lstlisting}

\lstset{
    basicstyle=\small\scriptsize,
    backgroundcolor=\color{gray!10}, 
    linewidth=\columnwidth,
    breaklines=true,
    frame=single, 
    rulecolor=\color{black}, 
    showstringspaces=false, 
    numbers=left, 
    numberstyle=\tiny\color{gray}, 
    xleftmargin=0em, 
    numbers=none, 
    moredelim=**[is][\color{gray!70}]{@}{@} 
}
\begin{lstlisting}[caption={Sample input for fine-tuning.}, label=lst:prompt2]
{
    "role": "system",
    "content": "### Task:
    You are given a sentence and a frame with its associated frame elements and sometimes examples. Your task is to label the frame elements in the sentence using JSON. Keys should only be one of the defined frame elements. Do not make up your own frame elements, and do not remove or change the input in any way. Identify the frame elements based on the highlighted target word. 
    
    ### Notes:
    - Return the tagged sentence in a ```json ``` code block.
    - Texts must not overlap."
},
{
    "role": "user",
    "content": "### Frame Information
    Frame Name: Law
    Frame Definition: A Law regulates activities or states of affairs within a Jurisdiction, dictating ... [omitted for brevity] ...

    Frame Elements:
    Law (Core): This FE identifies the rule designed to guide ... [omitted]
    ... [omitted]
    
    ### Input:
    Since the early 1990s , China has improved its export controls , including the promulgation of **regulations** on nuclear and nuclear dual - use exports and has pledged to halt exports of nuclear technology to un - safeguarded facilities."
},
{
    "role": "assistant",
    "content": "### Output: 
    ```json{'Law': 'regulations', 'Forbidden': 'on nuclear and nuclear dual - use exports'}```"
}
\end{lstlisting}

\begin{table}
    \centering
    \resizebox{\linewidth}{!}{
    \begin{tabular}{lcccc}
        \hline
        \textbf{Model} & \textbf{P} & \textbf{R} & \textbf{F1} & \textbf{Acc} \\ 
        \hline
        GPT-4o-mini     & 0.416 & 0.543 & 0.471 & 0.308 \\ 
        Deepseek V3     & 0.466 & \underline{0.665} & 0.548 & 0.377 \\ 
        GPT-4o          & \underline{0.550} & 0.642 & \underline{0.592} & \underline{0.420} \\
        \hline
        KID             & 0.741 & 0.773 & 0.756 & - \\
        AGED            & 0.757 & 0.776 & 0.767 & - \\
        \citet{Ai_Tu_2024} & 0.764 & 0.777 & 0.771 & - \\
        KAF-SPA & \textbf{0.819} & \textbf{0.807} & \textbf{0.813} & - \\
        \hline
    \end{tabular}}
    \caption{In-context learning performance comparison.}
    \label{tab:incontext_performance}
\end{table}


\subsection{Model Selection and Evaluation (RQ2)}
\label{sec:model-selection}
We evaluated the in-context learning performance with GPT-4o, GPT-4o-mini, and Deepseek V3 using the prompt in Listing~\ref{lst:prompt}. The results of these models are shown in Table~\ref{tab:incontext_performance}. These experiments included several exemplar sentences defined in each frame. Since these in-context learning methods use exemplar data, we include previous works which have used exemplar sentences.  
Additionally, we benchmark these LLM-based approaches against state-of-the-art systems, including KID~\cite{zheng-etal-2022-double}, AGED~\cite{aged2023}, and \citet{Ai_Tu_2024}.


\begin{table}
    \centering
    \resizebox{\linewidth}{!}{
    \begin{tabular}{lcccc}
        \hline
        \textbf{Model} & \textbf{P} & \textbf{R} & \textbf{F1} & \textbf{Acc} \\ 
        \hline
        Qwen 2.5-0.5B   & 0.716 & 0.682 & 0.699 & 0.537 \\
        Llama 3.2-3B    & 0.717 & 0.691 & 0.704 & 0.543 \\
        Llama 3.2-8B    & 0.736 & 0.711 & 0.724 & 0.567 \\
        Qwen 2.5-1.5B   & 0.748 & 0.719 & 0.733 & 0.579 \\
        Qwen 2.5-3B     & 0.765 & 0.740 & 0.752 & 0.603 \\
        Qwen 2.5-7B     & {0.769} & {0.754} & {0.762} & {0.615} \\
        GPT-4o-mini     & 0.774 & 0.762 & 0.768 & 0.624 \\
        Qwen 2.5-14B    & 0.782 & 0.772 & 0.777 & 0.635 \\
        Phi-4 (14B)     & \underline{0.793} & \underline{0.777} & \underline{0.785} & \underline{0.646} \\
        \hline
        Llama 3.3-70B   & 0.748 & 0.738 & 0.743 & 0.591 \\
        Qwen 2.5-32B    & 0.792 & {0.787} & {0.789} & {0.652} \\
        Qwen 2.5-72B    & \textbf{0.798} & \textbf{0.790} & \textbf{0.794} & \textbf{0.658} \\
        \hline
        \citet{lin-etal-2021-graph}     & -     & -     & 0.721 & - \\ 
        AGED                            & 0.750 & 0.752 & 0.751 & - \\ 
        KAF-SPA                         & \underline{0.760} & 0.743 & 0.751 & - \\ 
        \citet{Ai_Tu_2024}              & 0.756 & \underline{0.753} & \underline{0.755} & - \\ 
        \hline
    \end{tabular}}
    \caption{Performance of different models fine-tuned using JSON-exist. Models are ordered by F1 score and separated into size buckets 0-14B and 14B+.}
    \label{tab:finetune_performance}
\end{table}

For fine-tuning, we experimented with Llama 3.2 (3B, 8B), Llama 3.3 (70B), Qwen 2.5 (0.5B-72B), Phi-4 (14B), and GPT-4o-mini\footnote{Due to high training and inference costs, we did not fine-tune GPT-4o.}, as detailed in Table~\ref{tab:finetune_performance}. These models were fine-tuned exclusively on the full-text annotations without exemplar sentences. Consequently, we exclude methods that rely on exemplars for fine-tuning. Our fine-tuning prompt is shown in Listing~\ref{lst:prompt2}.

To assess the impact of instruction tuning, we compared the base and instruction-tuned variants of Qwen 2.5-7B. The instruction-tuned version performed significantly worse (0.703 vs. 0.768 F1 score), leading us to prioritize base models where available in subsequent experiments.

Our results show that Qwen 2.5 consistently outperforms Llama 3 across all model sizes. Most fine-tuned LLMs surpass previous state-of-the-art approaches, with Qwen 2.5 (3B) notably outperforming the much larger Llama 3.3 (70B). Among smaller-scale models, Phi-4 achieved the best performance, while at the larger scale, Qwen 2.5 (72B) outperformed all competitors, including the smaller models. Notably, these two LLMs surpassed the previous best-performing system~\citet{Ai_Tu_2024} by +3.0\% and +3.9\% F1 score, respectively.

\begin{table}
    \centering
    \begin{tabularx}{\linewidth}{Xcccc}
        \hline
        \textbf{Format} & \textbf{P} & \textbf{R} & \textbf{F1} & \textbf{Acc} \\ 
        \hline
        5 Most-FE & 0.605 & 0.699 & 0.649 & 0.480 \\ 
        5 Diverse & 0.648 & \underline{0.708} & 0.677 & 0.511 \\ 
        5 Random & \underline{0.717} & 0.675 & \underline{0.696} & \underline{0.533} \\ 
        \hline
        Full Dataset & \textbf{0.774} & \textbf{0.762} & \textbf{0.768} & \textbf{0.624} \\ 
        \hline
    \end{tabularx}
    \caption{Performance comparison of GPT-4o-mini fine-tuned on different partitions of the dataset.}
    \label{tab:finetune-data-subsample}
\end{table}


\subsection{Dataset Analysis}

\paragraph{Fine-tune Data Subsampling}
To reduce the costs associated with the high token count of the full training dataset, we first investigated whether strategic subsampling could reduce training overhead and cost while maintaining performance. We evaluated three distinct approaches: selecting up to five samples with the highest number of Frame Elements (5 Most-FE), randomly selecting up to five samples (5 Random), and selecting up to five samples that maximize Frame Element diversity (5 Diverse). Each of these approaches utilize approximately 15\% of the original training dataset. 

The results of this experiment, presented in Table~\ref{tab:finetune-data-subsample}, revealed an interesting trade-off. While the diversity-focused and FE-rich sampling strategies achieved higher recall, they resulted in lower F1 scores and precision compared to random sampling. This suggests that these targeted approaches enhanced the model's ability to identify a broader range of Frame Elements, but at the expense of precision on commonly occurring FEs. Because each of these approaches still fell significantly short of the full dataset's performance, we continue subsequent experiments with the entire dataset.

\begin{figure}
  \centering
  \includegraphics[width=\linewidth]{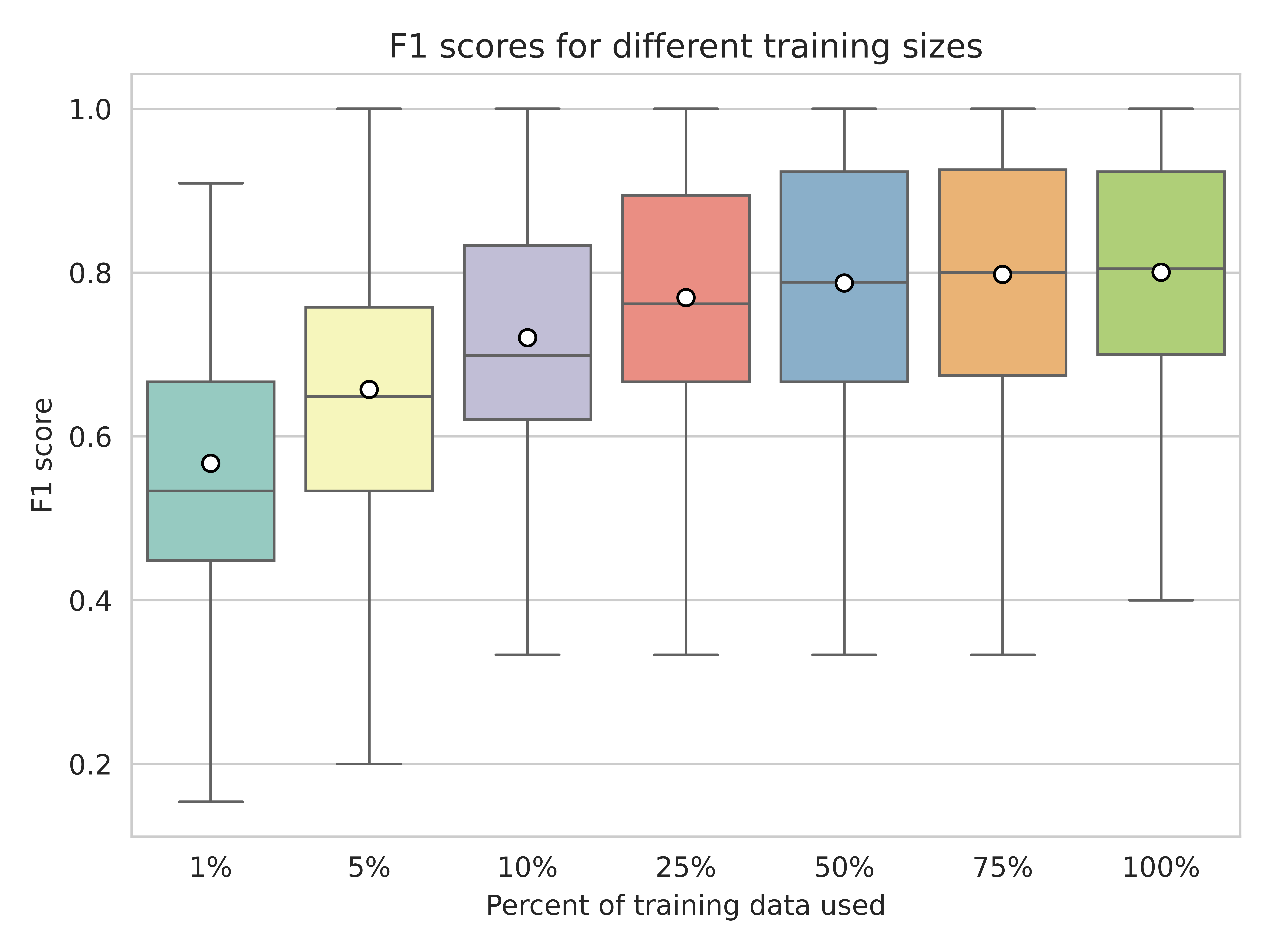}
  
  \caption{\label{fig:subsets-f1} Per-frame argument identification performance distribution for different training dataset sizes.}
\end{figure}
\begin{table}
    \centering
    \begin{tabular}{ccccc}
        \hline
        \textbf{Training \%} & \textbf{P} & \textbf{R} & \textbf{F1} & \textbf{Acc} \\ 
        \hline
        1\%                 & 0.551 & 0.471 & 0.508 & 0.340 \\  
        5\%                 & 0.652 & 0.590 & 0.619 & 0.448 \\  
        10\%                & 0.728 & 0.652 & 0.688 & 0.524 \\  
        25\%                & 0.767 & 0.726 & 0.746 & 0.595 \\  
        50\%                & 0.778 & 0.753 & 0.766 & 0.620 \\  
        75\%                & 0.781 & 0.776 & 0.779 & 0.638 \\  
        100\%               & 0.793 & 0.777 & 0.785 & 0.646 \\  
        \hline
    \end{tabular}
    \caption{Detailed per-frame argument identification performance distribution at different levels of data saturation.}
    \label{tab:saturation}
\end{table}
\paragraph{Data Saturation Analysis}
We also examined the relationship between training data volume and LLM performance through systematic experimentation with different dataset sizes during fine-tuning. We conducted this analysis using Phi-4, selected for its combination of strong performance and smaller model size. Each smaller subset is fully contained within larger ones to ensure consistency. 

The results of this analysis are presented visually in Figure~\ref{fig:subsets-f1} and in detail in Table~\ref{tab:saturation}. The performance trajectory shows distinct phases: a period of steady improvement from 1\% to 25\% of the dataset, followed by a transition to more modest gains beyond the 50\% mark. While the rate of average performance improvement diminishes after utilizing 50\% of the data, we observed two notable effects when using the complete dataset: a reduction in the inter-quartile range and enhanced performance on previously challenging frames. This suggests that additional training data continues to contribute to model robustness, even after average performance metrics begin to plateau.

\begin{table}
    \centering
    \begin{tabular}{lcccc}
        \hline
        \textbf{Format} & \textbf{P} & \textbf{R} & \textbf{F1} & \textbf{Acc} \\ 
        \hline
        All             & 0.793 & 0.777 & 0.785 & 0.646 \\ 
        Unseen Frame    & 0.725 & 0.691 & 0.708 & 0.548 \\ 
        Unseen FEs      & 0.560 & 0.477 & 0.515 & 0.347 \\ 
        \hline
    \end{tabular}
    \caption{Performance using a fine-tuned Phi-4 model on unseen samples.}
    \label{tab:unseen}
\end{table}
\subsection{Unseen and Out-of-domain Data (RQ3)}
\paragraph{Unseen Sample Evaluation}
We evaluate the ability of LLMs to identify frame elements on unseen and out-of-domain data in Table~\ref{tab:unseen}. We separate unseen data into two categories, Unseen (Frame) and Unseen (FEs). These categories correspond to test samples whose frames and frame elements are not seen in the training set, respectively. For this experiment we use a fine-tuned Phi-4 LLM for the same reason stated above.


Our analysis reveals a notable performance disparity between the two categories of unseen data. On unseen frames, where the entire frame is unseen in the training set, we observe a reduction in performance of -7.7\% F1 score compared to the overall performance. This relatively modest degradation suggests that the model has developed a robust general understanding of frame semantics that transfers reasonably well to new frames.

However, on unseen frame elements (FEs), we observe a substantially larger performance drop of -27.0\% F1 score. This significant degradation indicates a fundamental challenge in generalizing to entirely new frame elements. The disparity between these two scenarios provides valuable insights into the model's learning dynamics: the model appears to develop strong transferable knowledge about common frame elements that appear across multiple frames, enabling it to maintain reasonable performance even when encountering new frames that use familiar elements.

The stark performance difference with unseen FEs can be attributed to a few factors. First, unseen FEs are often highly specific to particular frames and may represent more nuanced or specialized semantic roles. Second, these elements typically have fewer analogous examples in the training data, limiting the model's ability to learn generalizable patterns. Third, the contextual cues for identifying these specialized FEs may be more subtle or require domain-specific knowledge that the model hasn't adequately acquired during training.

\begin{table}
    \centering
    \begin{tabularx}{\linewidth}{Xcccc}
        \hline
        \textbf{Model} & \textbf{P} & \textbf{R} & \textbf{F1} & \textbf{Acc} \\ 
        \hline
        GPT-4o           & 0.363 & 0.415 & 0.387 & 0.240 \\
        Phi-4            & 0.567 & 0.503 & 0.533 & 0.363 \\ 
        \hline
        SEMAFOR          & -     & -     & \textbf{0.570} & -     \\
        \hline
    \end{tabularx}
    \caption{Performance comparison on out-of-domain samples using the YAGS test set.}
    \label{tab:yags}
\end{table}
\paragraph{Out-of-domain Evaluation}
We also evaluate the performance of Phi-4 on out-of-domain samples using the YAGS dataset (Table~\ref{tab:yags}). We include an in-context learning GPT-4o implementation as a baseline along with SEMAFOR~\cite{das-etal-2014-frame}. SEMAFOR is one of the first frame-semantic parsing systems, and the only other previous work which was evaluated on the YAGS dataset; however, it is often outperformed by modern approaches. We found that both LLM implementations performed quite poorly, with GPT-4o achieving an F1 score of 0.387 and Phi-4 achieving 0.533. Both of these are lower than SEMAFOR's performance. This indicates an area where LLMs struggle significantly more than previous approaches. 

We also perform an assessment of the errors of these models to understand their cause. One observation we made is that there are many FEs in the YAGS dataset which are not defined in FrameNet. Another observation is that the sentences in YAGS tend to use poor grammar and often use slang. Additionally, we found that Phi-4's predictions were often more aligned with our human judgments than the original annotations, hinting at a possibility of data quality issues (further discussed in Appendix~\ref{app:yags-bad}). These factors, along with the new topics of discussion in the sentences are likely what leads to the poor performance on YAGS. 

\begin{table}
    \centering
    \resizebox{\linewidth}{!}{
    \begin{tabular}{lcccc}
        \hline
        \textbf{IFEval} & \textbf{BBH} & \textbf{GPQA} & \textbf{MUSR} & \textbf{MMLU-PRO} \\ 
        \hline
        -0.624 & 0.519 & 0.021 & \textbf{0.835} & 0.586 \\
        \hline
    \end{tabular}}
    \caption{Performance partial correlations computed for each benchmark.}
    \label{tab:benchmark-correlation}
\end{table}
\subsection{Benchmark Correlation Analysis}
Finally, we aim to understand what makes particular LLMs better than others on frame-semantic parsing. To do this, we analyze the correlation between our performance metrics and several common benchmarks for each LLM, where available. For this experiment, we focus on the IFEVal~\cite{zhou2023instructionfollowingevaluationlargelanguage}, BBH~\cite{suzgun2022challengingbigbenchtaskschainofthought}, GPQA~\cite{rein2023gpqagraduatelevelgoogleproofqa}, MUSR~\cite{sprague2024musrtestinglimitschainofthought}, and MMLU~\cite{hendrycks2021measuringmassivemultitasklanguage} benchmarks. We compute partial correlations between each benchmark and the F1 score on argument identification, accounting for model size as a confounding variable. Table~\ref{tab:benchmark-correlation} shows the correlation for each benchmark.

Our results indicate that MUSR exhibits the strongest positive correlation with frame-semantic parsing performance. Given that MUSR is designed to assess multistep reasoning, this suggests that models excelling in structured reasoning tasks also tend to perform well in frame-semantic parsing. Similarly, BBH and MMLU-PRO show strong positive correlations, aligning with their emphasis on complex reasoning and broad knowledge across multiple disciplines.

Interestingly, we observe a negative correlation with IFEval, which evaluates instruction-following capabilities using verifiable constraints. This suggests a potential trade-off between strict adherence to instructions and general problem-solving ability. This aligns with our earlier findings (Section~\ref{sec:model-selection}) that instruction-tuned models underperform their base versions on frame-semantic parsing. One possible explanation is that instruction-tuned models prioritize following explicit directives over deep semantic understanding.

\begin{table}
    \centering
    \begin{tabularx}{\linewidth}{Xcccc}
        \hline
        \textbf{Model} & \textbf{All} & \textbf{Amb} \\ 
        \hline
        Phi-4                           & 0.375 & 0.262 \\ 
        $\text{Phi-4}_{cand}$ w/o LF    & 0.882 & \textbf{0.862} \\ 
        $\text{Phi-4}_{cand}$ w/ LF     & 0.894 & \textbf{0.862} \\ 
        \hline
        KAF-SPA             & 0.912 & 0.776 \\
        KGFI                & 0.924 & 0.844 \\
        CoFFTEA             & \textbf{0.926} & 0.850 \\
        \hline
    \end{tabularx}
    \caption{Results on frame identification using frame element predictions.}
    \label{tab:candidate_frame}
\end{table}
\subsection{Frame Identification}
Previous work~\cite{devasier-etal-2024-robust} explored the possibility of filtering candidate targets produced by matching potential lexical units using a frame identification model. To build upon this idea towards a single-step frame-semantic parsing method, we explore the potential of frame elements being used to perform frame identification. In this approach, no ground-truth frame inputs are given. This also removes the bias from the model assuming the input always has at least one frame element. 

We compared this method with state-of-the-art approaches not using exemplar sentences, including KGFI~\cite{su-etal-2021-knowledge}, CoFFTEA~\cite{an-etal-2023-coarse}, and KAF-SPA~\cite{zhang2023knowledge}. We used Phi-4 for this experiment for the same reason as previous experiments. We found that directly using the model performed poorly, likely due to bias in the model's training using ground-truth frames, i.e., each input contains the given frame. To address this, we fine-tuned Phi-4 using candidates ($\text{Phi-4}_{cand}$) from the training set produced by \citet{devasier-etal-2024-robust} and achieved very strong performance. 

Sometimes frame elements are predicted for multiple candidate frames. When this happens, we randomly select one of the frames to be used as the prediction. Other options were explored, such as selecting the one with the most frame elements, only selecting the first frame, or utilizing GPT-4o as a tie-breaker, but none of these were effective. 

This method showed strong performance, particularly on ambiguous targets--targets with more than one possible frame-- where it achieved an accuracy of 0.862, higher than any previous approach. If we apply lexicon filtering~\cite{su-etal-2021-knowledge} on unambiguous targets, as is common among previous approaches, the overall accuracy is further increased to 89.4\%.






\section{Conclusion}


This work presents a comprehensive evaluation of large language models for frame-semantic parsing, with a particular focus on argument identification. Our systematic analysis reveals several important insights about the capabilities and limitations of LLMs in this domain. While LLMs demonstrate poor performance in zero-shot and few-shot settings, fine-tuned models achieve state-of-the-art results, with Qwen 2.5 (72B) surpassing previous approaches by a significant margin (+3.9\% F1 score).

Our investigation into input representations demonstrates that LLMs are sensitive to specific formats, with JSON-based formats achieving superior performance compared to alternatives. Our correlation analysis between frame-semantic parsing performance and common LLM benchmarks reveals that models excelling in multistep reasoning (as measured by MUSR) tend to perform better at argument identification, while instruction-following capabilities (measured by IFEval) show a negative correlation.

However, our results also highlight significant challenges. The substantial performance degradation on unseen frame elements (-27.0\% F1 score) and out-of-domain data indicates that current LLM approaches, despite their improvements over previous methods, still struggle with generalization. This limitation suggests that frame-semantic knowledge may not be sufficiently encoded, and that additional strategies may be needed to enhance model robustness across diverse contexts.

Our novel approach to frame identification using predicted frame elements shows promise, particularly for ambiguous targets, where it achieves state-of-the-art performance. This suggests that integrating frame element predictions into the frame identification process could be a valuable direction for future research.



\vspace{0.2cm}
\section*{Limitations}
Several methodological constraints impacted the scope and comprehensiveness of our analysis. Due to the substantial computational costs associated with fine-tuning large language models, we were unable to explore fine-tuning on certain high-performing models like GPT-4o and GPT-4. These models may have achieved even stronger results than those demonstrated in our current analysis.

Our experimental design relied on sequential parameter optimization to manage computational requirements. While this approach was practical, it introduces the possibility that certain combinations of parameters could yield unexpected results. For instance, XML representations might potentially outperform JSON embeddings when paired with 14B parameter models or applied to frame identification tasks. However, exploring these combinations was beyond the computational resources available for this study.

The scope of our research was limited to the English FrameNet dataset. As a result, our findings may not generalize to other languages or semantic frameworks. Cross-lingual validation would be necessary to establish the broader applicability of our approaches.

In the context of frame identification for ambiguous targets, our current method of handling multiple frames with predicted frame elements requires refinement. The randomized prediction approach can lead to inconsistent outputs. Additionally, our implementation used a fixed random seed of 0 for reproducibility, but we did not explore the potential impact of different random seeds on accuracy metrics.

Finally, our benchmark correlation analysis considered only model size as a confounding variable. This simplified approach may not account for other significant factors that could influence the relationship between benchmark performance and frame-semantic parsing capabilities. A more comprehensive analysis of confounding variables would provide deeper insights into these relationships.

\bibliography{custom,anthology}

\appendix


\section{Reproducibility}
\label{app:reproducibility}
To fine-tune the models in this work, we used three different systems. For the small models, (0.5-7B parameters) we experimented with training and evaluating on a system with 1x Nvidia RTX 4070 and another system with 1x Nvidia A100 40GB. For medium-sized models (14-32B parameters), we only experimented with the system with 1x Nvidia A100 40GB. For large models (70B+ parameters), we used a third system with 1x Nvidia H100 80GB. 

Github Copilot was used in the creation of some of our code.

\section{YAGS Quality Assessment}
\label{app:yags-bad}
\begin{table*}
\centering
\begin{tblr}{|Q[4cm, l]|Q[6cm, l]|Q[6cm, l]|}
\hline
\textbf{Sentence} & \textbf{YAGS Annotation} & \textbf{Our Annotation} \\
\hline
i feel that the pagan and wican be a lose people in **need** of a savior . & \{\textcolor{red}{'Dependent': 'at the pagan and wican be a lose people',} \newline
'Requirement': '\textcolor{red}{at the pagan and wican be a lose people in need} of a savior'\} & \{\textcolor{green}{'Cognizer': 'the pagan and wican',} \newline
'Requirement': 'of a savior'\} \\
\hline
how do u **get** rid of or cover up razor burn ? & \{'Entity': 'u'\} & \{'Entity': 'u', \newline 
\textcolor{green}{'Final\_quality': 'rid of or cover up razor burn'}\} \\
\hline

\end{tblr}
\caption{\label{tab:yags-bad-annotations} Examples of disagreements in our annotations compared to YAGS which may contribute to low performance.}
\end{table*}


We found several labels which we disagree with among a small random sample of Phi-4's predictions compared with the original annotations. We show these examples in Table~\ref{tab:yags-bad-annotations}.



\end{document}